\documentclass[sigconf]{acmart}

\usepackage{booktabs} 
\usepackage[utf8]{inputenc}
\usepackage{amsmath}
\usepackage{todonotes}
\usepackage[autolanguage]{numprint}
\usepackage{url}
\usepackage{listings,multicol}

\copyrightyear{2019}
\acmYear{2019}
\setcopyright{iw3c2w3}
\acmConference[WWW '19 Companion]{Companion Proceedings of the 2019 World Wide Web Conference}{May 13--17, 2019}{San Francisco, CA, USA}
\acmBooktitle{Companion Proceedings of the 2019 World Wide Web Conference (WWW '19 Companion), May 13--17, 2019, San Francisco, CA, USA}
\acmPrice{}
\acmDOI{10.1145/3308560.3316761}
\acmISBN{978-1-4503-6675-5/19/05}

\begin{document}
\title{Searching News Articles Using an Event Knowledge Graph Leveraged by Wikidata}

\author{Charlotte Rudnik}
\affiliation{%
  \institution{LIMSI, CNRS}
  \streetaddress{Rue du Belvedère}
  \city{Orsay}
  \country{France}
  \postcode{91406}
}
\email{charlotte.rudnik@limsi.fr}

\author{Thibault Ehrhart}
\affiliation{%
  \institution{EURECOM}
  \streetaddress{450 route des Chappes}
  \city{Sophia Antipolis}
  \country{France}
  \postcode{06410}
}
\email{thibault.ehrhart@eurecom.fr}

\author{Olivier Ferret}
\affiliation{%
  \institution{CEA, LIST,}
  \city{Gif-sur-Yvette, F-91191}
  \country{France}}
\email{olivier.ferret@cea.fr}

\author{Denis Teyssou}
\affiliation{%
  \institution{Agence France-Presse (AFP)}
  \city{Paris}
  \streetaddress{2, place de la Bourse}  
  \country{France}
  \postcode{75002}}
\email{denis.teyssou@afp.com}

\author{Raphaël Troncy}
\affiliation{%
  \institution{EURECOM}
  \streetaddress{450 route des Chappes}
  \city{Sophia Antipolis}
  \country{France}
  \postcode{06410}}
\email{raphael.troncy@eurecom.fr}

\author{Xavier Tannier}
\affiliation{
  \institution{Sorbonne Université, Inserm, LIMICS}
  \streetaddress{15, rue de l'école de médecine}
  \city{Paris}
  \postcode{75006}
  \country{France}
  }
\email{xavier.tannier@sorbonne-universite.fr}

\renewcommand{\shortauthors}{C. Rudnik et al.}

\begin{abstract}
News agencies produce thousands of multimedia stories describing events happening in the world that are either scheduled such as sports competitions, political summits and elections, or breaking events such as military conflicts, terrorist attacks, natural disasters, etc. When writing up those stories, journalists refer to contextual background and to compare with past similar events. However, searching for precise facts described in stories is hard. In this paper, we propose a general method that leverages the Wikidata knowledge base to produce semantic annotations of news articles. Next, we describe a semantic search engine that supports both keyword based search in news articles and structured data search providing filters for properties belonging to specific event schemas that are automatically inferred.
\end{abstract}

\maketitle


\section{Introduction}
\label{sec:introduction}
Information and communication technologies have provided tools and methods to make the production of information more democratic. In this context, journalists and technologists have developed the notion of ``data journalism'', which takes advantage of structured and numerical data in the production and distribution of news. It also takes advantage of the growing popularity of Linked and Open
Data and the development of structured knowledge bases such as DBpedia~\cite{Lehmann2013}, YAGO~\cite{Hoffart2013} or Wikidata~\cite{Vrandecic2014} to facilitate information analysis and to access a variety of points of view. However, knowledge is still far from being entirely represented in structured databases, and the most prominent way to convey information to the end user is still the free text, complemented by multimedia content. 

Intertwining structured and unstructured data in information systems is still an open research problem. In this paper, we present a system for aggregating unstructured news articles and structured data describing events leveraging on the Wikidata knowledge base.
This approach makes use of several Information Retrieval and Information Extraction tasks. In the context of Information Extraction, the knowledge associated with news articles can typically be used for training event extractors in a distant supervision mode~\cite{reschke-lrec-14}. From the Information Retrieval perspective, the approach makes it possible to retrieve news articles describing events using either keyword-based queries or filters that typically make use of properties available in knowledge bases. It also allows to query Wikidata and then to read an entire annotated article describing the corresponding event. We implemented a system which is available at \url{http://asrael.eurecom.fr/} and covers the last two  tasks.

\begin{figure*}
\centering
\includegraphics[width=0.75\linewidth]{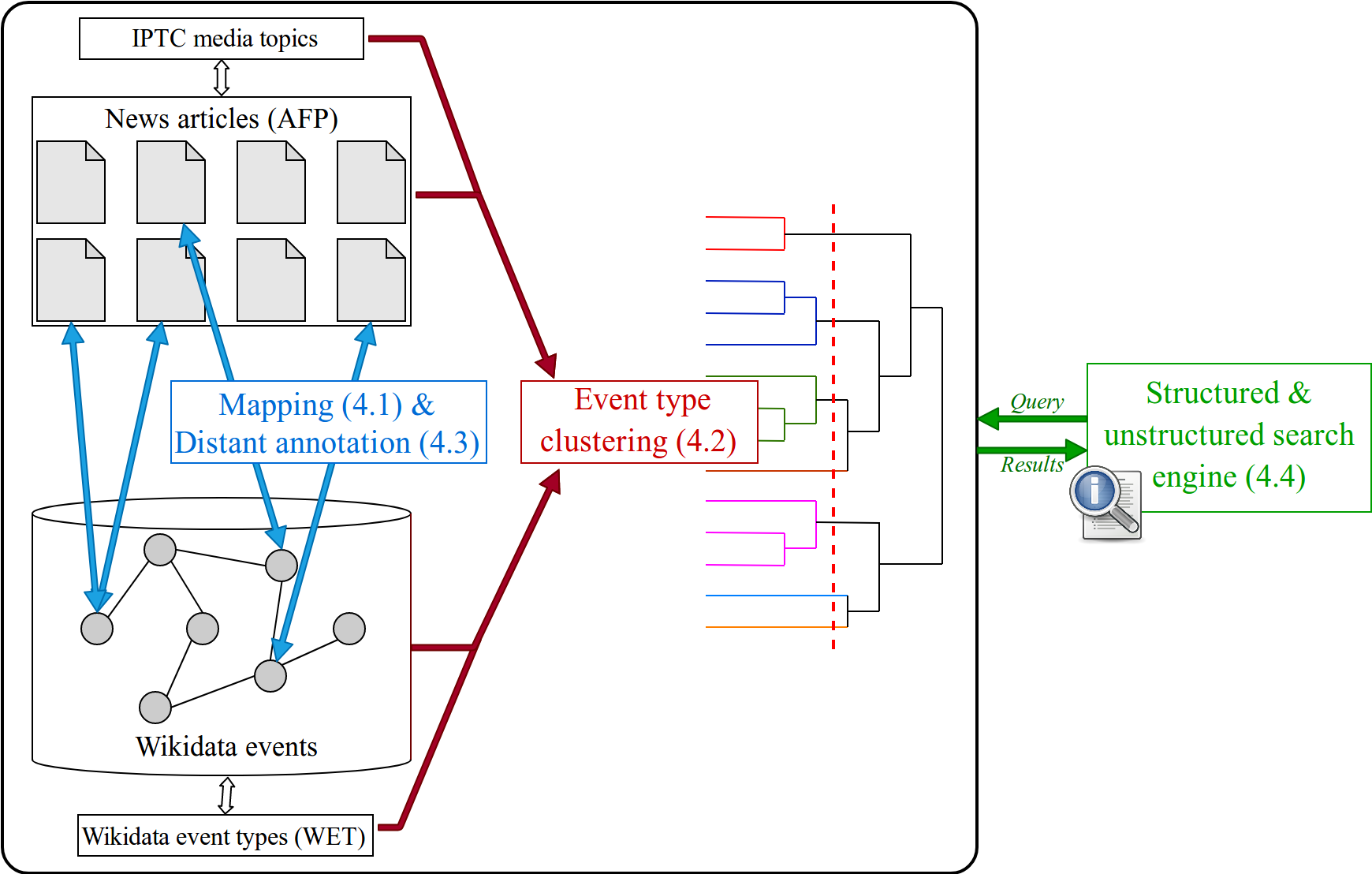}
\caption{System overview for annotating news articles and enabling structured search.} 
\label{fig:overview}
\end{figure*}

Figure~\ref{fig:overview} illustrates the architecture of our system. Events described in news articles are mapped to events from Wikidata (Section~\ref{sec:mapping}), and attributes from the Wikidata instances are used to annotate the news articles when possible (Section~\ref{sec:annotation}). Wikidata events belong to specific classes, but these classes are too fine-grained for being used in a search engine. Furthermore, many event classes actually share a similar structure (\emph{i.e.} sets of attributes). For example, a general \emph{election} schema is relevant for describing any type of elections regardless of the more specific Wikidata types such as ``\emph{Bundestag election}'' (\texttt{Q1007356}), ``\emph{direct election}'' (\texttt{Q1196727}), ``\emph{Elections in Saudi Arabia}'' (\texttt{Q4119635})\dots. For these reasons, we add a hierarchical clustering step (Section~\ref{sec:clustering}) to automatically create coarser grained schemas. Finally, we implemented an event-oriented knowledge graph and a search engine able to query and navigate through both the knowledge base and the news articles (Section~\ref{sec:search-engine}).

\section{Related Work}
\label{sec:related-work}
Contrarily to WikiNews\footnote{\url{https://www.wikinews.org/}}, Wikipedia does not aim to be a news service. However, Wikipedia's Current Events portal (WCEP\footnote{\url{https://en.wikipedia.org/wiki/Portal:Current_events}}) provides a set of pages where primarily events but also trends and developments are listed on a daily basis with links to reference articles. The WikiTimes project\footnote{\url{http://wikitimes.l3s.de/}} is the first attempt to build a structured and rich knowledge base of news events by harvesting the efforts of the Wikipedia crowd in maintaining WCEP~\cite{Tran2014}. The WikiTimes knowledge base is represented in RDF and it contains very short descriptions of events that can be filtered by entities, location and time.

Based on this experience, Gottschalk and Demidova have developed the EventsKG knowledge graph\footnote{\url{http://eventkg.l3s.uni-hannover.de/}}, a multilingual resource incorporating event-centric information extracted from Wikidata, DBpedia and YAGO, as well as less structured sources such as the Wikipedia Current Events Portal and Wikipedia event lists in five languages~\cite{gottschalk2018}. EventsKG re-uses the Simple Event Model (SEM) ontology~\cite{vanHage2011} to describe nearly \numprint{700000} events. However, temporal information is available for 76\% of those events and location information for only 12\% of them. While many entities are mentioned, extracted and disambiguated in the short descriptions of those events, this is far from being complete. Events are generally weakly categorized and both categorical and numerical data representing the events attributes are rarely extracted. Annotating semantically newsfeeds at scale is being continuously proposed in~\cite{trampus-is2012} which maintains the Newsfeed\footnote{\url{http://newsfeed.ijs.si/}} service. Annotations are, however, restricted to named entities that can be extracted by the Enrycher tool\footnote{\url{http://enrycher.ijs.si/}}. 

For  easing  the  exchange  of  news, the International Press Telecommunication Council (IPTC) has developed the NewsML Architecture (NAR), specialized into a number of languages such as NewsML G2 and EventsML G2. As part of this architecture, specific controlled vocabularies, such as the IPTC Media Topics or News Codes, are used to categorize news items together with other industry-standard
thesauri. In previous work, we designed an OWL ontology for the IPTC News Architecture and we converted the IPTC NewsCodes into a SKOS thesaurus~\cite{Troncy2008}. IPTC is now publishing itself the IPTC Media Topics in SKOS and has further developed the rNews vocabulary, largely based on Schema.org, for describing news articles. In this work, we re-use the rNews vocabulary to describe the original metadata attached to news articles. Furthermore, we annotate the news articles using properties and entities from Wikidata once events reported in the news have been mapped to existing Wikidata events.

While mapping text to knowledge has been the subject of a large body of work, represented in the recent ages with work such as \cite{mihalcea-cikm-07} or all the work about entity linking \cite{bunescu-eacl-06}, mapping text to event representations and more particularly news articles to event representations has not been the focus of lots of studies. One exception is \cite{mishra-pikm-14}, followed by \cite{mishra-ecir-16}, which tackles this kind of mapping according to an Information Retrieval perspective through two tasks based on the notion of \emph{Wiki-excerpt}. A \emph{Wiki-excerpt} corresponds to a description of an event built from Wikipedia and contains both a textual description and factual information about the event such as temporal expressions, geolocation and named entities. The first task, \emph{Wiki2News}, starts from a \emph{Wiki-excerpt} and aims at retrieving a set of past news articles about the considered event while \emph{News2Wiki} is the reverse task, consisting in retrieving  \emph{Wiki-excerpts} from a set of news articles. The work focuses more specifically on the \emph{Wiki2News} task by designing time-aware language models for supporting the retrieval of  past news articles. More recently, the \emph{Wiki2News} task of \cite{mishra-pikm-14} has been considered under the perspective of the enrichment of Wikipedia from a stream of news by \cite{lyu-wiki_workshop-18}. This work first builds a temporal event chain of the news articles related to the target event and then selects a subset of them according to various representativeness criteria exploited in a learning-to-rank framework. While all the work we have mentioned was based on Wikipedia, our work tackles the  \emph{Wiki2News} task by relying on Wikidata as a knowledge base, with a much simpler, still effective, approach.


\section{Data}
\label{sec:data}

\subsection{AFP News Articles}
In this work, we make use of a very large corpus of text newswire written in English provided by the French news agency AFP. More precisely, we use over 2~million articles covering the period 2004-2017. The topics are worldwide news ranging from politics, diplomacy, sports to natural disasters or economy and business. Each document is an XML file compliant with the NewsML standard, containing a title, a document creation time (DCT), a dateline where the article was written, one or several IPTC Media Topics and a set of keywords (slugs), as well as a textual content split into paragraphs.

The main topic of an article is generally a specific event, and sometimes, other older events are referred to in order to look at the current one from a wider perspective. This is why we consider that it is possible to associate an AFP article with one single event. Furthermore, we assume that the title and the first paragraph (lead) describe the event associated with the document. This is a realistic hypothesis since the basic rules of journalism impose that the first sentence should summarize the event by informing on the ``5~Ws'' (\emph{What}, \emph{Who}, \emph{When}, \emph{Where}, \emph{Why}). 

\subsection{Wikidata Occurrences}
Wikidata is a free and open knowledge base that can be read and edited by both humans and machines~\cite{Vrandecic2014}. It acts as a central storage for the structured data of its Wikimedia sister projects including Wikipedia, Wikivoyage, WikiNews, and others. In our work, we focus on newsworthy events, \emph{i.e.} something that happens within a locality and a temporality, and that could be described in one or more news articles. In regard to this definition, there are many general and specific classes in Wikidata that are related to events, but all these classes have in common the same parent class named ``Occurrence'' (\texttt{Q1190554}). For example, the event ``Cargolux Flight 7933'' (\texttt{Q3107014}) is an \textit{instance\_of} (\texttt{P31}) an ``aviation accident'' (\texttt{Q744913}) which is a \textit{subclass\_of} (\texttt{P279}) an ``aviation occurrence'' (\texttt{Q15733640}) which is a \textit{subclass\_of} (\texttt{P279}) ``occurrence'' (\texttt{Q1190554}).

In the remainder of this paper, we call Wikidata Event Type (or WET) the value of the property \emph{instance\_of} (\texttt{P31}) of a Wikidata event. In the previous example, ``aviation accident'' (\texttt{Q744913}) is a WET. An instance can have several Wikidata Event Types. 

We lined up on the temporal coverage of the AFP articles corpus and we considered all Wikidata event instances during the period from 2004 to 2017. The event date can be represented by three properties in Wikidata:
\begin{itemize}
  \item \texttt{P585}:\emph{point\_in\_time} if the event is a one-off event;
  \item \texttt{P580}:\emph{start\_time} and \texttt{P582}:\emph{end\_time} if the event has a duration.
\end{itemize}
In the second case, we consider the events that have started and ended during the period 2004-2017.

\begin{figure}
 \centering
 \includegraphics[width=\linewidth]{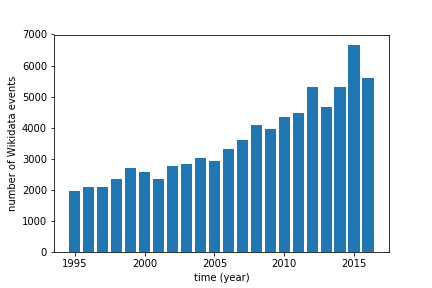}
 \caption{Evolution of the number of Wikidata events over time.} 
 \label{fig:wikiEvolution}
\end{figure}

The selection of Wikidata events consists of 60k Wikidata instances. As shown in Figure~\ref{fig:wikiEvolution}, the distribution over the years is not uniform. The increase in the last few years is explained by a better quality of the data and particularly by the presence of a date in the description of the events, but should not be interpreted as an increase of the number of events happening in the world.

\subsection{IPTC Media Topics}
\label{sec:iptc}
\begin{figure*}
 \centering
 \includegraphics[width=0.85\textwidth]{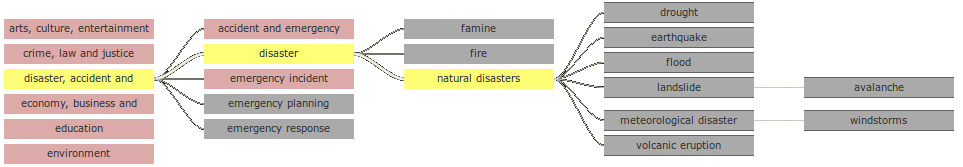}
 \caption{Excerpt of the IPTC Media Topics taxonomy.} 
 \label{fig:iptc}
\end{figure*}

The International Press Telecommunications Council (IPTC) maintains a taxonomy of Media Topics\footnote{\url{http://cv.iptc.org/newscodes/mediatopic}, \url{http://show.newscodes.org/index.html?newscodes=medtop}}, which can be seen as a controlled and hierarchical set of indexing keywords. Each article written by AFP is associated with at least one IPTC code by its author. IPTC Media Topics (later called IMTs) give information about the topic of the article and are often linked with a type of event (earthquake, election, crash\dots), but not always (politics, theatre\dots). Figure~\ref{fig:iptc} illustrates a part of the IPTC Media Topic hierarchy.

\section{Approach}
\label{sec:approach}
Following the description of the data in Section~\ref{sec:data}, we define that ``AFP article'' stands for the main event associated with the AFP article and described in the lead, while ``Wikidata event'' stands for the structured events described in our selection of Wikidata instances. The term 'article' will refer to the news article whereas the term instance will refer to a Wikidata instance. 

\subsection{Mapping AFP with Wikidata}
\label{sec:mapping}
In order to map Wikidata events to AFP articles, we consider that two mentions of the same event share the following characteristics: same time, same place and same type or category (election, natural disaster, etc.), which we defined as a content similarity.

\subsubsection{Scoring function}
For mapping a Wikidata instance to an AFP article, we define the following criteria: 
\begin{description}
  \item [Date.] The article has to be written at most the day following the \emph{point\_in\_time}, or between \emph{start\_time} and \emph{end\_time} if the event has a duration.
  \item [Location.] The location of a Wikidata instance is defined by the properties \textit{country} (\texttt{P17}) and \textit{location} (\texttt{P276}). One of these values must have be mentioned in the AFP article. 
  \item [Subject.] The Wikidata Event Type (WET) and the title of the instance define a list of keywords relevant to the subject of the article. The similarity score is then the sum of a IMT-based \emph{tf.idf} of the keywords occurring also in the AFP article:
  \begin{equation}
     score(article) = \sum_{t_i \in article} {tf.idf_{IMT}(t_i) \times \mathbf{1}_{keywords} (t_i)} 
  \end{equation}
where 
    $$\mathbf{1}_{keywords} (t_i) = \left\{
                    \begin{array}{ll}
                      1 & \text{if $t_i \in WET$} \\
                      0 & \text{otherwise}\\
                    \end{array}
                  \right.$$
and 
\begin{equation}
 \begin{aligned}
tf.idf_{IMT}(t_i) &= tf_{IMT}(t_i) \times log \frac{N}{df_{IMT}(t_i)} 
 \end{aligned}
\end{equation}
with $t_i$ the $i^{th}$ token in the article, $df_{IMT}(t_i)$ the number of IMTs associated with documents containing $t_i$, $tf_{IMT}(t_i)$ the number of occurrences of $t_i$ in the articles sharing the same IMT, and $N$ the number of IMTs. If the AFP article has more than one IMT, then the highest $tf_{IMT}$ for each word is considered. 

The weights are based on IPTC Media Topics. Hence, they increase the relevance of a token in a particular context. For instance, ``Police'' will be more relevant in an event regarding a crime compared to an event regarding an earthquake, even if the token is present in both articles.

An event is considered to be about the same subject as a Wikidata instance if this score is higher than a threshold. Empirically, this threshold is set to 0.04 and roughly corresponds to two tokens in common between the article and the list of keywords.
\end{description}

\subsubsection{Evaluation}
After this processing, \numprint{97160}~articles have been mapped to \numprint{8350}~Wikidata instances from \numprint{933}~Wikidata event types (out of 42k Wikidata events in the considered temporal range). Note that it is neither necessary nor desirable that all articles be mapped to Wikidata instances. Most AFP articles do not relate to events that are meant to end up in Wikidata (\emph{i.e.}, a political speech, a reaction to an event or a trivial news event) and a lot of Wikidata events are not described by any press agency (book publication, local festival or cultural event, TED talks\dots).

In order to evaluate the quality of this mapping, we manually built a set of \numprint{406}~pairs \emph{(article, Wikidata~instance)} based on \numprint{88}~instances explicitly linked to a WikiNews page\footnote{\url{https://en.wikinews.org}} in Wikidata. 
Only a very small part of Wikidata events are linked to a WikiNews page\footnote{Note that our automatic mapping could be used to automate this Wikidata/WikiNews mapping.} but we used them to ensure that only ``media-worthy'' Wikidata events are considered in our manual evaluation. We share this gold standard for further comparison\footnote{\url{https://github.com/crudnik/asrael}}.

We report in Table~\ref{tab:mapping_results} the precision (correct mapping rate), recall ($1 - $missed mapping rate) and F1-measure, when considering the first~3, the first~5 or all the sentences in the AFP articles. These scores show the high quality of the mapping, and consequently, only a very tiny fraction of incorrect information will be shown to the user. Note that missed mappings do not prevent the event to be queried and visualized by our search engine. We only miss the link between the structured and the unstructured data.

\begin{table}
\centering
\begin{tabular}{|c|c|c|c|}
\hline
\textbf{\# sentences} & Precision & Recall    & F1-score \\
\hline
3                     & \numprint{1.00} &  \numprint{0.67}  &  \numprint{0.80} \\
5                     & \numprint{0.99} &  \numprint{0.71}  &  \numprint{0.83} \\
all                   & \numprint{0.96} &  \numprint{0.75}  &  \numprint{0.84} \\
\hline
\end{tabular}
\caption{\label{tab:mapping_results} Evaluation of the mapping between AFP articles and Wikidata events.}
\end{table}


\subsection{Schema Clustering}
\label{sec:clustering}
As depicted in Table~\ref{table:schemaWetMap}, Wikidata event types are often fine-grained and the subclass hierarchy can vary in terms of quality and depth. We seek more coarse-grained event categories for the relevance and robustness of our news classification and potential filters for our search engine. Indeed, from a human perspective, several WETs (e.g. NATO summits and G20 summits) share the same or a very similar structure and clustering them together will make the classification process easier, as well as simplify the interaction with the user within the search engine interface.

\begin{table}
  \centering
  \setlength{\tabcolsep}{3pt}
  \renewcommand{\arraystretch}{1.3}
  \begin{tabular}{ |l|p{5.5cm}| }
    \hline
    \textbf{Target schema} & \textbf{Related Wikidata event types} \\ \hline
    Election & Bundestag election, direct election, \newline Elections in Saudi Arabia, ... \\
    Plane crash & Aviation accident, plane crash, \newline mid-air collision, ... \\
    Summit & NATO summit, G20 summit, ... \\
    \hline
  \end{tabular}
\caption{Examples of Wikidata Event Types for three target schemas.}
\label{table:schemaWetMap}
\end{table}

To do so, we adopted a hierarchical clustering method based on 3~similarity measures. Each similarity is based on a different representation of the Wikidata Event Type.
\begin{itemize}
  \item \textbf{Label representation}: Even if composed of only a few words (see Table~\ref{table:schemaWetMap}), the labels of the Wikidata Event Type (WET) can be a good clue for deciding whether two clusters are similar or not. For instance, the labels ``\textit{Election in UK}'', ``\textit{Election in France}'' and ``\textit{Election}'' should be clustered together. However, this approach needs to exclude mentions of organizations or locations, as in ``\emph{earthquake in New Zealand}'' compared to ``\emph{New Zealand general election}'', which are arguably not similar in terms of event type. Therefore, we replace mentions of named entities with generic tokens using the spaCy\footnote{\url{http://spacy.io/}} named entity recognition system following the Table~\ref{tab:genericization} (e.g. ``\emph{France}'' is replaced by ``\emph{\textsc{geopolitical entity}}'', ``\emph{jan-7}'' by ``\emph{\textsc{date}}'). 

\begin{table}
\centering
\begin{tabular}{|c|c|}
\hline
\textbf{NER label} & \textbf{New mentions} \\[2pt]\hline
GPE & geopolitical entity \\
ORG & organization \\
PERSON & person \\
NORP & nationality \\
DATE & date \\
\hline
\end{tabular}
\caption{\label{tab:genericization} Generic tokens.}
\end{table}

The representation of the labels $R_l(T)$ of a WET $T$ is the mean of the word2vec~\cite{Mikolov2013} vectors of their words.
\begin{equation}
 R_{l}(T) = \mathop{mean}_{w \in{label(T)}}(w2v(w))
\end{equation}
where $label(T)$ is the set of words in $T$'s label and $w2v(w)$ is the word2vec representation of the word $w$. 

  \item \textbf{Content representation}: this representation is based on the content of the articles. A \emph{WET document} is built by concatenating all the articles mapped to this WET at the previous step (Section~\ref{sec:mapping}). The content representation $R_{c}(T)$ is then a vector of all words $t_i$ weighted by their $tf.idf_{WET}$, computed as follows: 
\begin{equation}
 tf.idf_{WET}(t_i) = tf_{WET}(t_i) \times log \frac{M}{df_{WET}(t_i)}
\end{equation}
where $tf_{WET}(t_i)$ is the number of occurrences of the term in the WET document, $df_{WET}(t_i)$ is the number of WET documents containing the term, and $M$ is the total number of $WETs$ in our dataset.

  \item \textbf{IPTC Media Topic representation}: in order to improve and to facilitate the stopping decision of the clustering, we add a feature based on the IMTs (see Section~\ref{sec:iptc}). 


As each AFP article is associated with one or several IMTs, the mapping described in Section~\ref{sec:mapping} provides also a mapping between a WET and a list of IMTs. We interpret these codes as a new vocabulary describing the WET and we use again a \emph{tf.idf} representation of this new vocabulary. The representation $R_{imt}(T)$ is a sparse vector of size equal to the total number of IMTs present in the corpus, where, for each IMT $imt$: 

\begin{equation}
\begin{aligned}
tf.idf_{T}(imt) = tf_{T}(imt) \times log \frac{M}{df_{T}(imt)}
\end{aligned}
\end{equation}

where $tf_{T}(imt)$ is the number of articles labeled by the IMT $imt$ that have been mapped to a Wikidata event of type~$T$, $df_{T}(imt)$ is the number of WETs mapped with at least one article with label $imt$, $M$ is the total number of $WETs$ in our dataset. 
\end{itemize}

We use these three representations to compute the following similarity between two WETs $T_i$ and $T_j$:
\begin{align}
sim(T_i,T_j) &= \alpha \times \cos(R_{l}(T_i), R_{l}(T_j)) \\
&+ \beta \times \cos(R_{c}(T_i), R_{c}(T_j)) \\
&+ \gamma \times \cos(R_{imt}(T_i), R_{imt}(T_j))
\end{align}

where $\cos$ is the cosine similarity measure and the weights $\alpha$, $\beta$ and~$\gamma$ are empirically set as $\numprint{0.38}$, $\numprint{0.57}$, $\numprint{0.05}$ respectively. 

Note that each of the ``label'', ``content'' and ``IMT'' representations has a different role. Increasing $\alpha$ gives a higher weight to very short texts, which are generally difficult to compare~\cite{Lee2016}. This would, for example, make closer labels such as `strike', `general strike' and `military strike', which would decrease the quality of the clustering. Content representation is based on longer pieces of text, but increasing $\beta$ would give a higher weight to potential errors in the mapping. 
Finally, the IMT similarity is quite categorical compared to the other ones. We want to use it only as an help for choosing when to stop the clustering, which explains the low value of~$\gamma$ in the global similarity score. 
The importance of this balance between all sources of information is at the same time a strength and a limitation of the method.

\begin{figure}
 \centering
 \includegraphics[width=\linewidth]{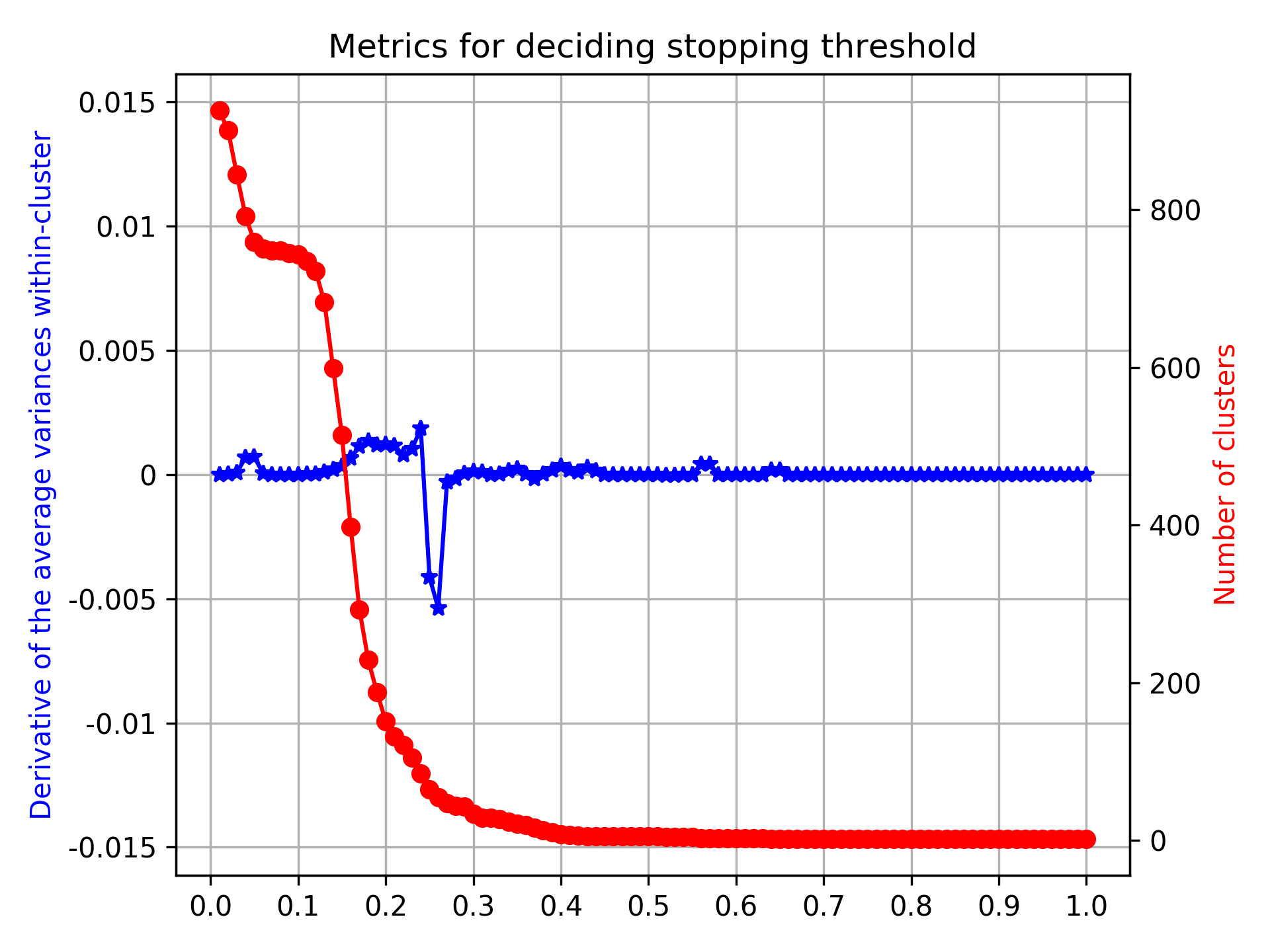}
 \caption{Number of clusters and derivative of the distance within clusters as a function of the similarity threshold to cut the dendrogram.}
 \label{fig:Elbow}
\end{figure}

\begin{figure}
 \centering
 \includegraphics[width=\linewidth]{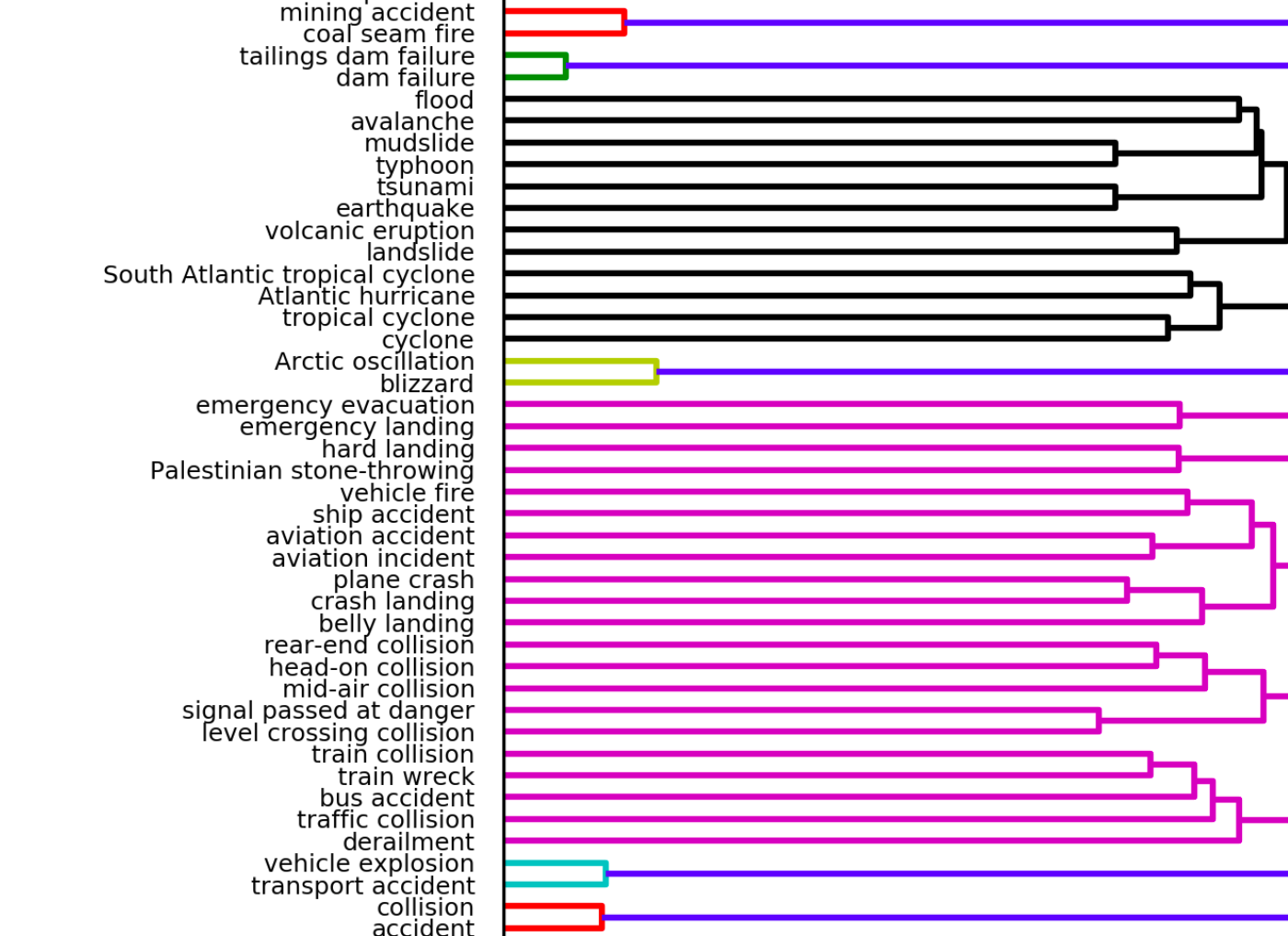}
 \dots \\
 \includegraphics[width=\linewidth]{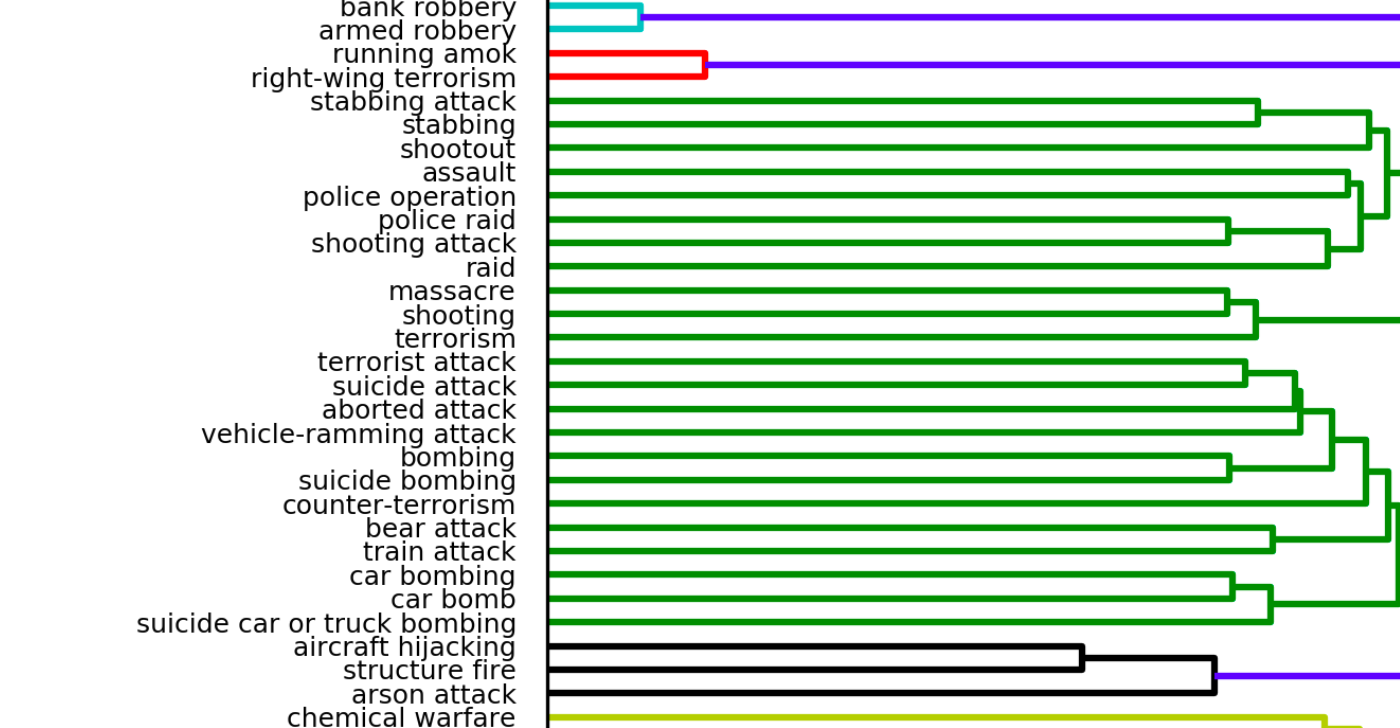}
 \dots \\
 \includegraphics[width=\linewidth]{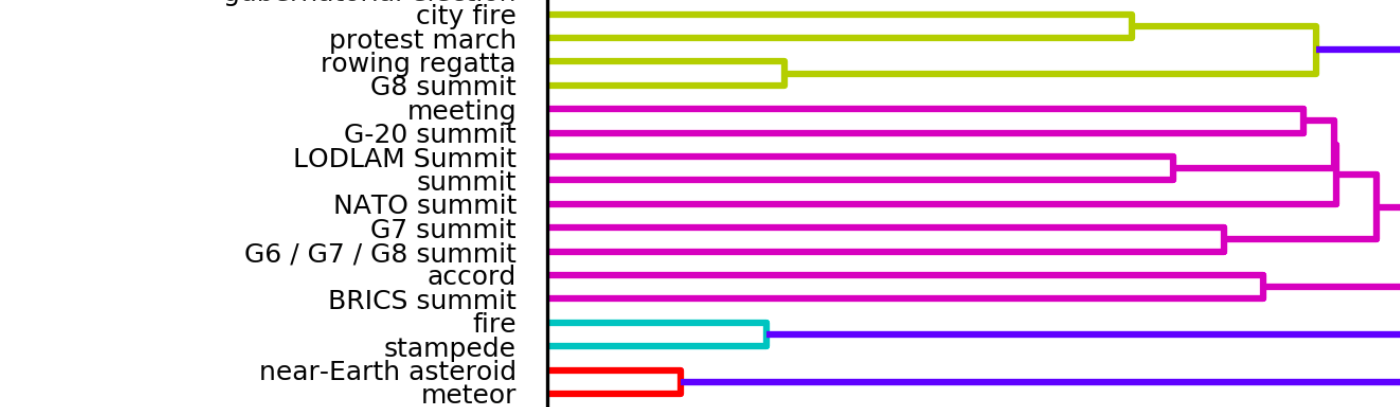}
 \dots \\
 \includegraphics[width=\linewidth]{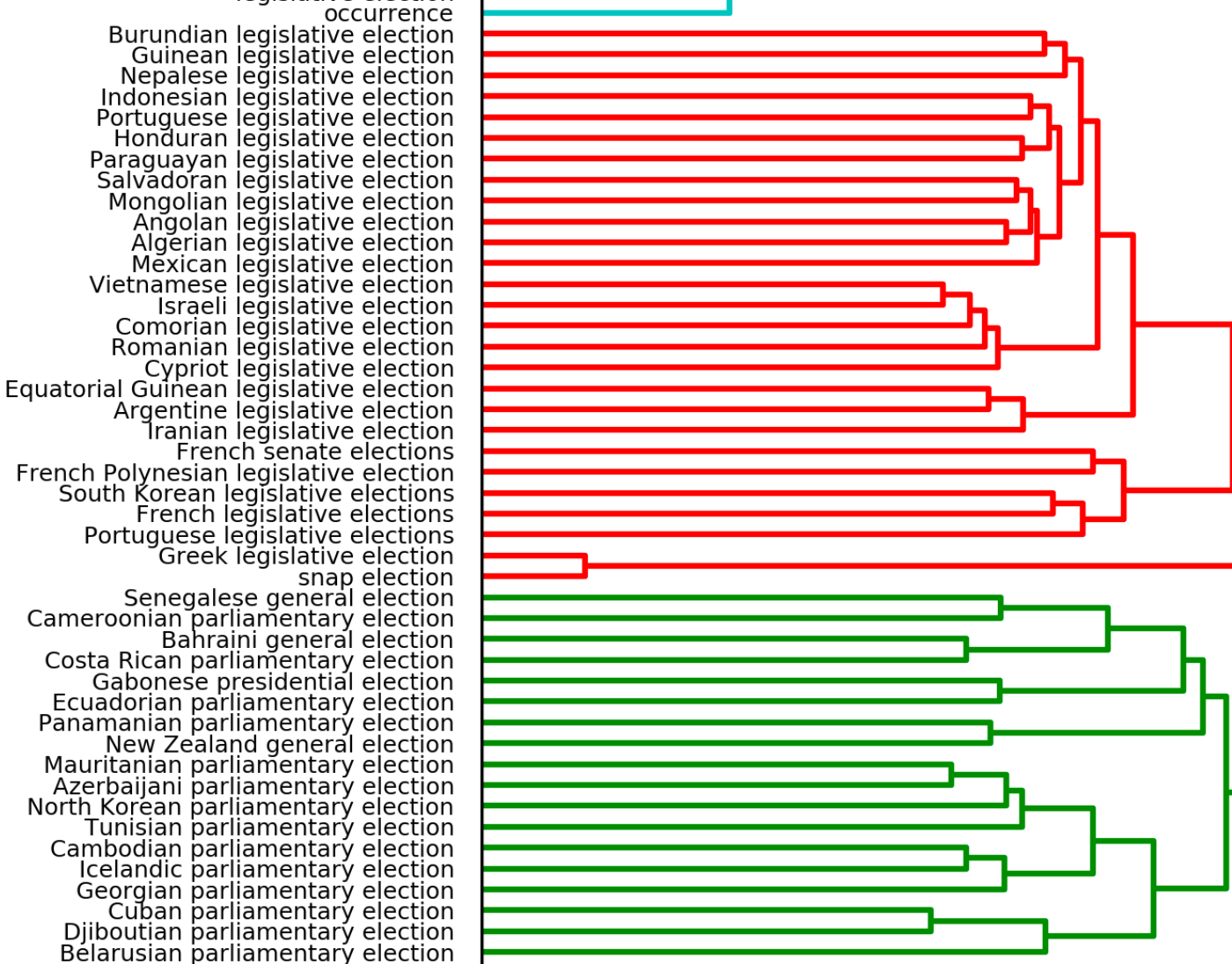}
 \caption{\label{fig:dendrogram} Extracts of the dendrogram issued after clustering.}
\end{figure}

Our agglomerative hierarchical clustering procedure is based on the Ward's method. To cut the resulting dendrogram we used a threshold defined by the elbow method, aiming at finding at which number of clusters the marginal gain of variance will start dropping. According to the graph of Figure~\ref{fig:Elbow}, the best threshold is \numprint{0.23}.
This leads from \numprint{933} initial Wikidata event types to \numprint{119} clusters in total. Some extracts of the dendrogram are available in Figure~\ref{fig:dendrogram}.

We observe that this clustering step enables to group together natural disasters, or summits (NATO, G8, ...) into coherent clusters. We also obtained several Election clusters, with the three main ones which seem related to legislative elections, parliamentary elections and general elections.

We empirically evaluated the quality of the clusters for choosing the parameters of our model. Building a protocol for a formal evaluation of this step is a future work. As for most clustering tasks, there is no unique good solution and an automatic, reproducible evaluation seems difficult to set up. 


\subsection{Automatic Semantic Annotation of News Articles}
\label{sec:annotation}
Our objective is to annotate semantically AFP news articles leveraging Wikidata structured data describing events being told in those articles. We distinguish Wikidata properties that have textual values from the ones having numerical values.

\subsubsection{Entity Annotation}
When Wikidata properties have a textual value, our goal is to search whether this value is present or not in the news article. We also use DBpedia redirections to collect different variations of the mention to search in the text.

\subsubsection{Quantity Annotation}
When Wikidata properties have a numerical quantity, the problem of finding this information in the news article is much harder. The news articles come as a continuous stream and some reported information generally evolves over time, such as the death toll after a deadly accident. Consequently, a proper semantic annotation of a news article should not always consider the exact quantity value indicated in Wikidata. Therefore, for quantitative values, we introduce some flexibility and a quantity in the text (float or integer) is added to a candidate list of annotations if it is in the +/- 10\% range of a property and if it is in the first five sentences of the article (supposedly, the earlier in the text, the most relevant to the article main subject). These candidates are then ranked according to the relevance of the semantic context of the numerical value in the text (considering that the context expresses the type of the attribute). Consequently, given two quantities, the more the context (e.g. ``9 were killed on Saturday'') is similar with the property type (e.g. \emph{number\_of\_killed} rather than \emph{magnitude\_on\_Richter\_scale}), the more probable the quantity is to be linked with this property. The news article is then annotated with the wikidata property and the most relevant quantity.

\subsubsection{Serializing the Annotations}
We represent both the news articles metadata and the semantic annotations in RDF. We first convert the AFP news article metadata encoded in NewsML in RDF using the rNews vocabulary. For example, the metadata associated with the news article described by Listing~\ref{lst:rnews}
indicates that this story was created on 24/03/2015 with the English headline \emph{'No survivors' in Germanwings crash: transport minister}. 

\begin{minipage}{\linewidth}
\begin{lstlisting}[captionpos=b, caption=Semantic annotation of a news article, label=lst:rnews, basicstyle=\tiny\ttfamily,breaklines=true]
<http://asrael.eurecom.fr/news/71e6c1b5-cbfa-3f85-8510-e200652f6735>
  a  rnews:Article ;
  rnews:dateCreated "2015-03-24T12:41:21Z"^^xsd:dateTime;
  rnews:headline  "'No survivors' in Germanwings crash: transport minister"@en ;
  dc:subject  <http://cv.iptc.org/newscodes/subjectcode/03013000>,
    <http://cv.iptc.org/newscodes/subjectcode/04015000>,
    <http://cv.iptc.org/newscodes/subjectcode/03010000>, 
    <http://cv.iptc.org/newscodes/subjectcode/04000000>,
    <http://cv.iptc.org/newscodes/subjectcode/03010003>, 
    <http://cv.iptc.org/newscodes/subjectcode/04015001>, 
    <http://cv.iptc.org/newscodes/subjectcode/03000000>;
  schema:keywords   "minister", "aviation", "accident", 
     "Germany", "Spain", "survivors", "France" .
\end{lstlisting}
\end{minipage}

The event being described in this news article exists in Wikidata as \texttt{Q19671417}. 
We create an instance of the \texttt{schema:Event}\footnote{The \texttt{schema} prefix refers to the Schema.org vocabulary.} which is about this news article. In this article, the number of dead people (150) is correctly found (Listing~\ref{lst:annotation}). The schema \emph{S34} is one of the schema output of the clustering phase described in the section~\ref{sec:clustering}.

\begin{minipage}{\linewidth}
\begin{lstlisting}[captionpos=b, caption=Semantic annotation of a news article, label=lst:annotation, basicstyle=\tiny\ttfamily, breaklines=true]
<http://asrael.eurecom.fr/news/71e6c1b5-cbfa-3f85-8510-e200652f6735>
   rnews:about  <http://asrael.eurecom.fr/event/71e6c1b5-cbfa-3f85-8510-e200652f6735> .

<http://asrael.eurecom.fr/event/71e6c1b5-cbfa-3f85-8510-e200652f6735>
   a    schema:Event , wd:Q750215 , rnews:Concept ;
   rdfs:label  "'No survivors' in Germanwings crash: transport minister" ;
   dc:identifier "urn:newsml:afp.com:20150324T124135Z:TX-PAR-ENS90:5" ;
   owl:sameAs wd:Q19671417 ;
   wdt:P1120  "150" ;
   wdt:schema "S34" .
\end{lstlisting}
\end{minipage}

\subsubsection{Annotation Dataset}
As a result, we created annotations associated with \numprint{370} properties extracted from Wikidata. This dataset can easily be used for a relation extraction task, with a distant supervision system. The evaluation of this automatic annotation is part of our future work. Note that this step is not necessary to build the search engine described hereafter.

\subsection{Search Engine}
\label{sec:search-engine}
\begin{figure*}
 \centering
 \includegraphics[width=0.75\linewidth]{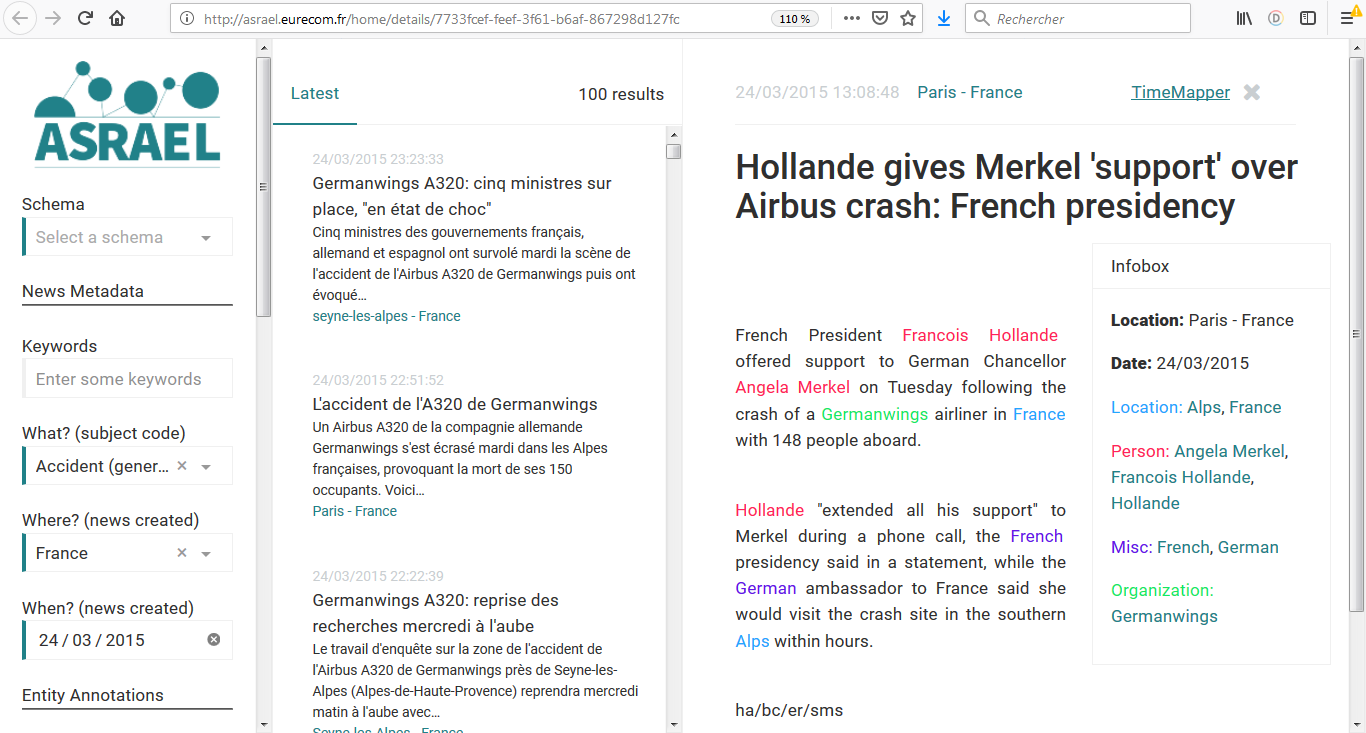}
 \caption{Search engine filtering articles describing plane crash events occurring in France on 24 March 2015.}
 \label{fig:asrael-NER}
\end{figure*}

\begin{figure*}
 \centering
 \includegraphics[width=0.75\linewidth]{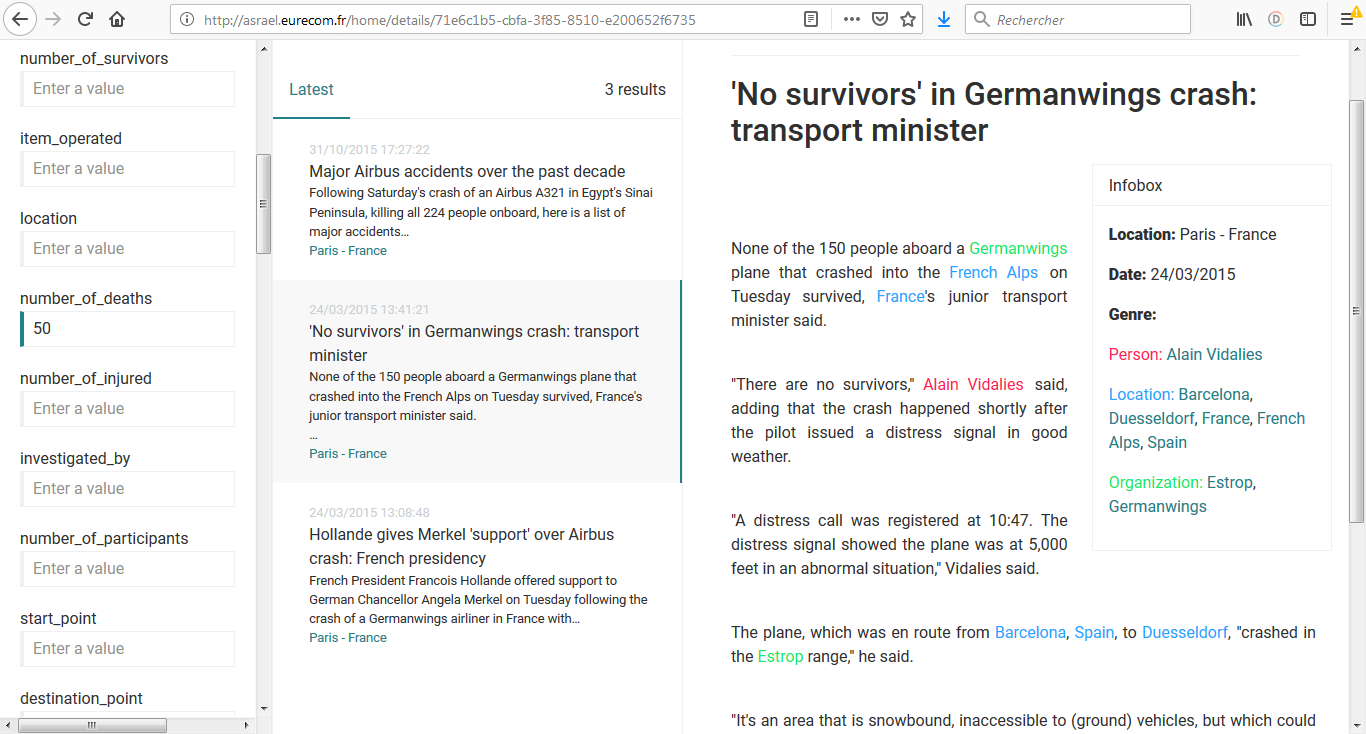}
 \caption{Search engine filtering articles describing plane crash events having caused at least 50 victims.}
 \label{fig:asrael-schema}
\end{figure*}

We load all RDF annotations in a triple store using the Openlink Virtuoso software. The full text of the news article is also indexed in the triple store. We then developed a user interface that performs SPARQL queries to provide views on the data.
The Figure~\ref{fig:asrael-NER} depicts the view of the news article\footnote{\tiny\url{http://asrael.eurecom.fr/home/details/7733fcef-feef-3f61-b6af-867298d127fc}}. On the top right, we show an infobox composed of the main named entities extracted in the article using the ADEL system~\cite{plu:SWJ2019}.
The Figure~\ref{fig:asrael-schema} depicts the view of the news article\footnote{\tiny\url{http://asrael.eurecom.fr/home/details/71e6c1b5-cbfa-3f85-8510-e200652f6735}}. On the left panel, the user has selected the schema \emph{S34} corresponding to crash accident. Therefore, a set of additional properties are automatically suggested as new filters, such as the number of victims. The user has entered the value 50 and the search engines retrieves the news articles that describe crash accidents that have yielded at least this number of victims.

\section{Conclusion and Future Work}
\label{sec:conclusion}
In this paper, we develop an event-based search engine capable to query both the structure data of knowledge bases and the unstructured textual content of news articles. This facilitates the navigation through events of the same type and aggregate complementary information about the same event. Furthermore, we produced a semi-automatically annotated text dataset. This dataset could be used as a distant supervision for training an annotation system. This system could then be able to extract the structure of the events from the news article, even if they are not in Wikidata. We also plan to work on a multilingual support for this system.

\section*{Acknowledgements}
This work has been partially supported by the French National Research
Agency (ANR) within the ASRAEL Project, under grant number
ANR-15-CE23-0018, and the ContentCheck Project, under grant number ANR-15-CE23-0025-01.

\bibliographystyle{ACM-Reference-Format}


\end{document}